# Physics-informed ConvNet: Learning Physical Field from a Shallow Neural Network

Pengpeng Shi[1,2,a], Zhi Zeng[1,b], Tianshou Liang [c]

[a] School of Civil Engineering, Xi'an University of Architecture and Technology, Xi'an 710055, Shaanxi, China

[b] School of Mechano-Electronic Engineering, Xidian University, Xi'an 710071, Shaanxi, China

[c] School of Mechanical and Electrical Engineering, Xi'an University of Architecture and Technology, Xi'an 710055, Shaanxi, China

**Abstract:**

Big-data-based artificial intelligence (AI) supports profound evolution in almost all of science and technology. However, modeling and forecasting multi-physical systems remain a challenge due to unavoidable data scarcity and noise. Improving the generalization ability of neural networks by "teaching" domain knowledge and developing a new generation of models combined with the physical laws have become promising areas of machine learning research. Different from "deep" fully-connected neural networks embedded with physical information (PINN), a novel shallow framework named physics-informed convolutional network (PICN) is recommended from a CNN perspective, in which the physical field is generated by a deconvolution layer and a single convolution layer. The difference fields forming the physical operator are constructed using the pre-trained shallow convolution layer. An efficient linear interpolation network calculates the loss function involving boundary conditions and the physical constraints in irregular geometry domains. The effectiveness of the current development is illustrated through some numerical cases involving the solving (and estimation) of nonlinear physical operator equations and recovering physical information from noisy observations. Its potential advantage in approximating physical fields with multi-frequency components indicates that PICN may become an alternative neural network solver in physics-informed machine learning.

---

[1] The authors have contributed equally to this paper.
[2] Corresponding author. shipengpeng@xjtu.edu.cn (P. Shi)
E-mail addresses: shipengpeng@xjtu.edu.cn (P. Shi), zhizeng@mail.xidian.edu.cn (Z. Zeng), liangtianshou@gmail.com (T. Liang).

# 1. Introduction

As a branch of artificial intelligence, machine learning uses algorithms to identify essential information and abstract structures in data and create models for prediction [1]. Neural network is an important part of machine learning algorithms. It is a computer program inspired by the biological neural network, which establishes a high-dimensional nonlinear empirical relationship between input and output variables by simulating the way the human brain processes information [2]. As an efficient data-driven model, artificial neural networks have been widely used in smart cities [3], intelligent machines [4], and other fields [5]–[7]. In convention, big data with high quality is a key to the successful application of artificial neural networks [8]. However, field data acquisition and large-scale testing involve unaffordable costs for some complex physical and engineering technologies, let alone the noise often contained in data [9]. In such cases, the convergence of the training process cannot even be guaranteed. This unavoidable data scarcity and noise becomes the main obstacle to deploying artificial neural networks in complex physical disciplines and engineering technology [10].

An effective solution is to embed existing domain knowledge to the network, and the construction of a new generation of machine learning models combined with physical law has become a research hotspot [11]–[15]. Estimating the physical field described by differential equations with a few boundary data can be regarded as a typical problem with small sample size. Recently, the physics-informed neural network (PINN) has been considered a promising numerical tool that uses a fully connected neural network (FCN) to approximate the physical field governed by differential equations [16]–[18], which has been shown to be applicable to various mathematical and physical problems [19]–[24]. In the method, the neural network act as a high-dimensional regressor, and the loss comes from the combination of boundary conditions and governing equations. The physical field is estimated iteratively by updating the network weights through the gradient descent method. PINN is a meshless algorithm that can effectively integrate observation data from different sources and is even effective for partially understood problems with uncertain high-dimensional observations [25]. It is also efficient and accessible for uncertain large-scale inverse problems [20]. Although PINN has potential for complex physical problems, the spectral bias nature of FCN may directly limit its capability in multi-scale problems with complex high-frequency components [26]. Essentially, the deep series-parallel connection of continuous function families with undetermined parameters in FCN is inefficient in approximating the nonlinear physical field when both high and low-frequency components coexist [27].

Developing a robust and efficient framework is the core for the next generation of physical knowledge-based machine learning. Shallow nested convolutional neural networks are promising in dealing with the spectral bias problem in PINN because they can avoid exacerbating network complexity by reducing width and depth when fitting the nonlinear physical field [28]–[30]. Based on this consideration, this paper proposes a novel physics-informed convolutional network (PICN for short) to respond to the spectral bias problems encountered by PINN. In the proposed method, the physical field is generated by a deconvolution layer and a single convolution layer. Differentials field with their combinations is constructed using the pre-trained shallow convolution operators. An efficient linear interpolation network calculates the loss function involving boundary conditions and the physical constraints in irregular geometry domains. This method can be regarded as a highly simplified upgrading of existing generative networks [31] with the operator neural networks [32][33], whose convergence quality is much better than that of the PINN with an actual calculation speed of more than an order of magnitude.

The remainder of this paper is organized as follows. Section 2 introduces the PICN network with detailed implementations. Section 3 discusses its convergence properties in the frequency domain. Sections 4-6 present numerical cases of solving various nonlinear partial differential equations, estimating the differential equations of physical field, reconstructing and denoising display of physical field from noisy observations. A large number of numerical examples verify the effectiveness of the method. Finally, some remarks and conclusions are given in the last section.

## 2. PICN with detailed implementations

### 2.1. Problem statement

The overall goal of this work is to develop a novel artificial neural network with physical information incorporated. The construction and training of networks enriched by mathematical-physical knowledge provide a new paradigm in physical modeling and computation with limited data, e.g., a few noisy measurements from an experiment. The physical law governing the physical field compensates the data scarcity and low quality, which is usually described by parameterized nonlinear partial differential equations as [16]

$$\frac{\partial u}{\partial t} + \mathbf{N}[u;\lambda] = 0, x \in \Omega, t \in [0,T] \tag{1}$$

where $u(t,x)$ denotes the solution of physical field, $\mathbf{N}[u;\lambda]$ is a nonlinear differential operator with $\lambda$ as its parameter, and $\Omega$ is a subset of $\mathbf{R}^D$. Physics-informed machine learning aims to infer all the physical states of interest in the spatiotemporal domain $u(t,x)$ from physical information like initial conditions, boundary conditions, and measurement data. This work focuses on two cases: (I) data-driven solution with physical differential equations from limited boundary observation; (II) data-driven discovery of physical field from noisy observations. For the first problem, the physical information is described by a simplified nonlinear partial differential equation

$$\frac{\partial u}{\partial t} + \mathbf{N}[u] = 0, x \in \Omega, t \in [0,T] \tag{2}$$

Recent physics-informed machine learning represents a new framework for solving partial differential equations based on artificial intelligence. Solving partial differential equations, including initial and boundary conditions, corresponds to training neural networks.

## 2.2. Structure of PICN

The recently reported PINN's architecture has been almost exclusively based on the fully connected feed-forward neural network where the differential field is obtained by auto-differentiation [34]. In comparison, the proposed PICN's architecture benefits from the convolutional neural network where the differential field is obtained from frozen pre-trained filters. The diagram describes the PICN training process.

Let us consider the data-driven solution for a scalar physical field governed by a two-dimensional nonlinear partial differential equation:

$$\mathbf{T}(x,y,u,u_x,u_y,u_{xx},u_{xy},u_{yy},\ldots) = 0 \quad \mathbf{x} \in \Omega \tag{3}$$

$$\mathbf{B}[u(\mathbf{x})] = f(\mathbf{x}), \quad \mathbf{x} \in \Gamma \tag{4}$$

where $\mathbf{T}$ is an arbitrary operator, and $u$ is the physical field, a function of variable $\mathbf{x}$. In addition, $\mathbf{B}$ is an operator related to the boundary conditions. In the current study, the following Neumann and Dirichlet boundary conditions for the physical field are considered

$$\begin{cases} u(\mathbf{x}) = f_{\Gamma_1}(\mathbf{x}) & \mathbf{x} \in \Gamma_1 \\ \dfrac{\partial u(\mathbf{x})}{\partial \mathbf{n}} = \nabla u(\mathbf{x}) \cdot \mathbf{n}(\mathbf{x}) = f_{\Gamma_2}(\mathbf{x}) & \mathbf{x} \in \Gamma_2 \end{cases} \quad (5)$$

in which $\mathbf{x} = [x, y]^T$, $\Gamma$ represents the boundary, and $\mathbf{n}$ represents the normal vector to the boundary.

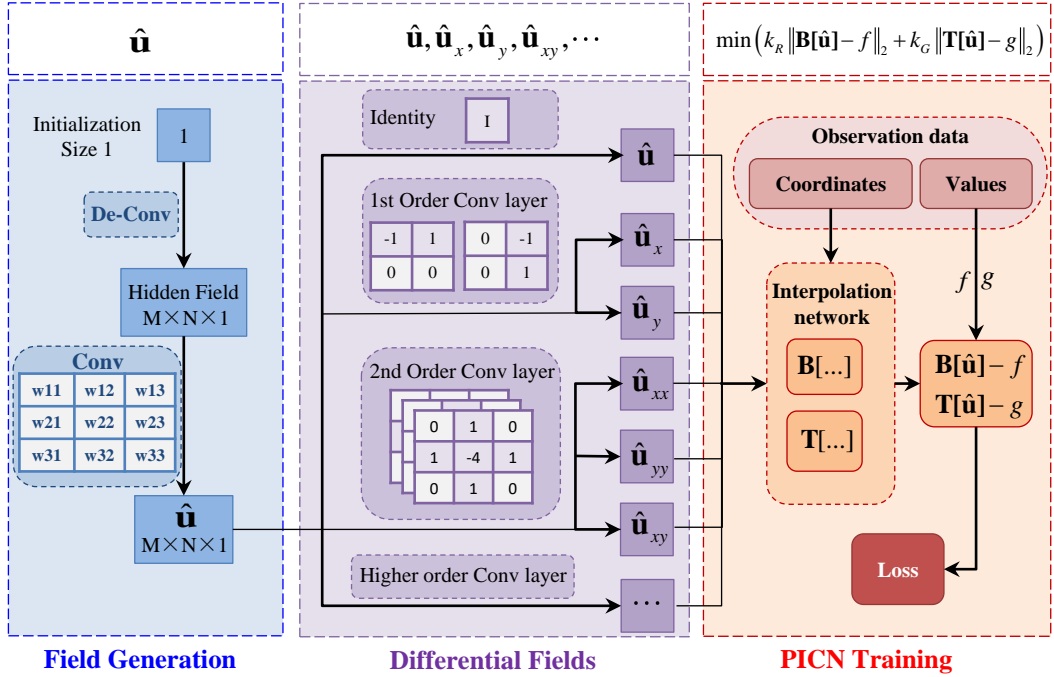

**Fig. 1** The schematic diagram of PICN. Due to the advantages of shallow architecture, the proposed PICN neural network accelerates the convergence speed and improves the convergence quality. The network structure consists of three parts: 1) a convolutional neural network for physical field generation; 2) a pre-trained convolutional layer to calculate differential fields of the estimated physical field; 3) an interpolation network for loss analysis in irregular geometry domains.

The proposed PICN method is shown in Figure 1. It consists of three parts. In the ***first*** part, a deconvolutional neural network is used to generate an estimate of a hidden field from a constant input value. The estimation can be formulated as

$$\mathbf{y}^h = \mathbf{w}^h x^{in} + b^h \in \mathbb{R}^{m \times n} \quad (6)$$

where $x^{in}=1$, $\mathbf{w}^h=\left[w_{i,j}^h\right]\in\mathbb{R}^{m\times n}$ is the matrix of weights, and $b^h\in\mathbb{R}$ is the vector of biases corresponding to the hidden layer. The hidden field is then passed through a convolutional layer to produce an estimated physical field. That is

$$\hat{\mathbf{u}}=a\left(\mathbf{y}^h*\mathbf{w}^o+b^o\right)\in\mathbb{R}^{m\times n} \tag{7}$$

in which $*$ stands for the convolution operation, $\mathbf{w}^o=\left[w_{i,j}^o\right]\in\mathbb{R}^{p\times q}$, $b^o\in\mathbb{R}$ is the bias, $a(\cdot)$ is the activation function, the output $\hat{\mathbf{u}}$ is an $m\times n$ matrix. Entries in $\hat{\mathbf{u}}$ represent the value of the physics at regular discrete grid points.

In the **second** part, CNNs with pre-trained weights are used to extract differential fields from the estimated physical field. The classical finite difference method demonstrates the effectiveness of calculating derivatives of the physical field on discretized grids. The numerical format of the finite difference can be obtained strictly from the Taylor series expansion [35]. Taking the Laplace operation as an example, the corresponding CNN kernel $\mathbf{W}_\Delta$ is

$$\Delta u=\frac{\left[u(x+\Delta,y)+u(x-\Delta,y)+u(x,y+\Delta)+u(x,y-\Delta)\right]-4u(x,y)}{\Delta^2}+o\left(\Delta^2\right) \tag{8}$$

The Laplace operation can be calculated as

$$\Delta\hat{u}(x_i,y_j)=(\hat{\mathbf{u}}*\mathbf{w}_\Delta)_{i,j},\ \mathbf{W}_\Delta=\frac{1}{\Delta^2}\begin{bmatrix}0&1&0\\1&-4&1\\0&1&0\end{bmatrix} \tag{9}$$

In the **third** part, the estimation of the physical field with its various derivatives is combined according to governing equations and boundary conditions. All parameters of the networks can be learned by minimizing the following sum of the squared errors loss function.

$$L=k_R\left(L_{R_1}+L_{R_2}\right)+k_G L_G \tag{10}$$

where

$$L_G = \frac{1}{N_\Omega} \sum_{\mathbf{x} \in \Omega} \left[ \mathbf{T}\left(x, y, \hat{u}, \hat{u}_x, \hat{u}_y, \hat{u}_{xx}, \hat{u}_{xy}, \hat{u}_{yy}, \ldots \right) \right]^2$$

$$L_{R_1} = \frac{1}{N_{\Gamma_1}} \sum_{\mathbf{x} \in \Gamma_1} \left[ \hat{u}(\mathbf{x}) - f_\Gamma(\mathbf{x}) \right]^2 \quad (11)$$

$$L_{R_2} = \frac{1}{N_{\Gamma_2}} \sum_{\mathbf{x} \in \Gamma_2} \left[ \nabla \hat{u}(\mathbf{x}) \cdot \mathbf{n}(\mathbf{x}) - f_N(\mathbf{x}) \right]^2$$

where $\mathbf{x} = \{\mathbf{x}_i\}$ are the considered positions of physical information data, including boundary conditions and measurement data. $L_{R_1}$ and $L_{R_2}$ represent the losses corresponding to the two types of boundary conditions, $L_G$ represents the loss corresponding to the governing equation of the physics field. The final neural network loss function is constructed by a linear (or nonlinear) combination of these loss functions.

It is worth noting that the position of the sample may not be on the regular grid. Therefore, we need to build an interpolation network to estimate the field values that are not on the grid. Obviously, a regular grid can be constructed whose region covers the positions of all training samples. Here, a bilinear interpolation network is proposed as

$$\hat{u}(x, y) = \frac{1}{\Delta^2} \begin{pmatrix} \hat{u}(\underline{x}, \underline{y})(\overline{x} - x)(\overline{y} - y) \\ +\hat{u}(\underline{x}, \overline{y})(\overline{x} - x)(y - \underline{y}) \\ +\hat{u}(\overline{x}, \underline{y})(x - \underline{x})(\overline{y} - y) \\ +\hat{u}(\overline{x}, \overline{y})(x - \underline{x})(y - \underline{y}) \end{pmatrix} + O(\Delta^2) \quad (12)$$

where $\underline{x}$ represents the abscissa of the nearest node whose abscissa is not greater than $x$. $\overline{x}$ represents the abscissa of the nearest node whose abscissa is larger than $x$. The definitions of $\underline{y}$ and $\overline{y}$ are similar. The derivative field can also be interpolated using this bilinear interpolation network. For each sampling location, the interpolation constants can be pre-computed in advance in the network. For smooth and uniformly bounded fields, the interpolation error tends to zero when the mesh is sufficiently dense.

## 2.3. Data-driven discovery of physical law

Besides solving equations as shown in Sec. 2.2, another important application of PICN is the discovery of partial differential operators from measurement data of a physical field. In other words, the task becomes to find the optimal parameter $\lambda$ in the governing equation using the measurement data. This paper provides two approaches for the discovery of data-driven physical law. 1) Freeze the filters in the second part of the PICN network and replace the nonlinear operator **T** with a trainable interpolation network; 2) Filters in the second part of the PICN network are also participated in the training. The following numerical examples will demonstrate the effectiveness of this method. In addition, subsequent research confirms that the above method could also be used for denoising the display of physical information from noisy observations.

## 3. Discussion on the convergence behavior

### 3.1. Convergence analysis in the frequency domain

The spectral bias of FCN has been discussed in previous studies [26]. Since PINN adopts FCN as its backbone, it also inherited the spectral bias of FCN in the training process. In comparison, PICN is promising for approximating physical fields with multi-frequency components by taking advantage of the shallow convolutional network architecture. For convenience, the one-dimensional case is discussed here.

Consider first the physics generation part of PICN. This part consists of a deconvolution layer and a convolution layer. The output of this section can be expressed as:

$$\hat{u}_j = \sigma\left(\sum_{u=(1-P)/2}^{(P-1)/2} \left(w^{(1)}_{j-u} \cdot 1 + b^{(1)}\right) w^{(2)}_{(P+1)/2-u} + b^{(2)}\right) \tag{13}$$

in which $P$ indicates the size of the convolutional kernel. Rewriting the above formula into continuous form obtains

$$\hat{u}(x) = \sigma\left(\int_{-\delta}^{+\delta} w^{(2)}(s)\left(w^{(1)}(x+s) + b^{(1)}\right) ds + b^{(2)}\right) \tag{14}$$

where $\delta$ indicates the size of the convolutional kernel. If one extends the domain of $w^{(2)}(s)$ to infinite and assumes that the values outside its original domain are 0, the above formula can be further changed to

$$\hat{u}(x) = \sigma\left(\int_{-\infty}^{+\infty} w^{(2)}(s)\left(w^{(1)}(x+s) + b^{(1)}\right) ds + b^{(2)}\right) \tag{15}$$

The equation can be shortened as

$$\hat{u}(x) = \sigma\left[\Upsilon(x) + b^{(2)}\right] \tag{16}$$

where $\Upsilon(x) = w^{(2)}(x) * \overline{w}^{(1)}(x)$, $\overline{w}^{(1)} = w^{(1)}(x) + b^{(1)}$.

For physical field observations, the network loss can be expressed as

$$L_{R_1} = \frac{1}{N}\sqrt{\sum_{x_i}\left(\hat{u}(x_i) - u(x_i)\right)^2} \tag{17}$$

Since $\sigma$ is an invertible function at low frequencies, the minimization of the above loss function is equivalent to minimizing the following function:

$$L'_{R_1} = \frac{1}{2}\sum_{x_i}\left(\Upsilon(x_i) - g(x_i)\right)^2 \tag{18}$$

where $g(x) = \sigma^{-1}[u(x)] - b^{(2)}$.

According to Parseval's theorem, the discrete Fourier transform of this function can be written as

$$F'_{R_1}(\omega) = \frac{1}{2}|D(\omega)|^2 \tag{19}$$

in which $D(\omega) \triangleq F[\Upsilon(x)](\omega) - F[g(x)](\omega)$. In above equation $F[u(x)](\omega)$ is the Fourier transform of $u(x)$, and then

$$F[\Upsilon(x)](\omega) = F[w^{(2)}(x)](\omega) F[\overline{w}^{(1)}(x)](\omega) \tag{20}$$

in which

$$F[w^{(2)}(x)](\omega) \approx \sum_{n=0}^{N-1} w^{(2)}(n)\exp\left(-i\frac{2\pi}{N}\omega n\right) \tag{21}$$

Therefore, we have

$$\frac{\partial D(\omega)}{\partial w^{(2)}_s} = \exp\left(-i\frac{2\pi}{N}\omega n\right) F[\overline{w}^{(1)}(x)](\omega) \tag{22}$$

Let

$$D(\omega) = A(\omega) e^{i\theta(\omega)} \tag{23}$$

Then, one has

$$\begin{aligned}
\frac{\partial F'_{R_1}(\omega)}{\partial w^{(2)}_s} &= \overline{D(\omega)}\frac{\partial D(\omega)}{\partial w^{(2)}_s} + D(\omega)\frac{\partial \overline{D(\omega)}}{\partial w^{(2)}_s} \\
&= A(\omega)\left[\begin{array}{l}\exp(-i\theta(\omega))\exp\left(-i\frac{2\pi}{N}\omega n\right)F\left[\overline{w}^{(1)}(x)\right] \\ +\exp(i\theta(\omega))\exp\left(i\frac{2\pi}{N}\omega n\right)\overline{F\left[\overline{w}^{(1)}(x)\right]}\end{array}\right](\omega) \\
&= A(\omega)\exp(i\theta(\omega))\exp\left(i\frac{2\pi}{N}\omega n\right) \\
&\quad \left[\exp(-2i\theta(\omega))\exp\left(-i\frac{4\pi}{N}\omega n\right)\frac{F\left[\overline{w}^{(1)}(x)\right]}{\overline{F\left[\overline{w}^{(1)}(x)\right]}}+1\right]\overline{F\left[\overline{w}^{(1)}(x)\right]}(\omega)
\end{aligned} \tag{24}$$

According to Parseval Theorem, one has $\dfrac{\partial L'_{R_1}}{\partial w^{(2)}_j} = \sum_\omega \dfrac{\partial F'_{R_1}(\omega)}{\partial w^{(2)}_j}$. Thus, one obtains

$$\left|\frac{\partial L'_{R_1}}{\partial w^{(2)}_j}\right| \leq 2\left|A(\omega)W^{(1)}(\omega)\right| \tag{25}$$

As can be seen from these two equations, $A(\omega)$ is bounded. Thus the gradient does not change significantly with increasing frequency. The training parameters $w^{(1)}$ in the deconvolutional layer can be discussed similarly. Therefore, the PICN method does not have a frequency domain bias for the loss function corresponding to the physical field observations.

Next, let us analyze the loss $L_G$. Without loss of generality, it is assumed here that there is only one *k*-order derivative term in this loss. The minimization of the above loss function is equivalent to minimizing the following function:

$$L'_G = \frac{1}{2}\left(\frac{\partial^k \Upsilon(x)}{\partial x^k} - \frac{\partial^k g(x)}{\partial x^k}\right)^2 \tag{26}$$

the discrete Fourier transform of this function can be written as:

$$L'_G = \sum_\omega F'_G(\omega) \tag{27}$$

According to Parseval's theorem, one has

$$F'_G(\omega) = \frac{1}{2}|D(\omega)|^2 \tag{28}$$

where

$$D(\omega) \triangleq F\left[\frac{\partial^k \Upsilon(x)}{\partial x^k}\right](\omega) - F\left[\frac{\partial^k g(x)}{\partial x^k}\right](\omega) \tag{29}$$

in which

$$\begin{aligned}F\left[\frac{\partial^k \Upsilon(x)}{\partial x^k}\right](\omega) &= (i\omega)^k F[\Upsilon(x)](\omega) \\ &= (i\omega)^k F[w^{(2)}(x)](\omega) F[\overline{w}^{(1)}(x)](\omega)\end{aligned} \tag{30}$$

As signal frequency increases, the magnitude of the nominal gradient increases polynomially. Therefore, for the PICN method with the participation of the differential governing equations, the high-frequency components in the physical field are preferentially learned.

### 3.2. Comparison of and discussion on training convergence

This subsection compares the training process when using PINN and PICN to solve some differential equations. Firstly, a one-dimensional nonlinear differential equation containing trigonometric functions and a square operator is analyzed. The following kinetic process along with initial value conditions is considered,

$$\begin{aligned}\sin(u^2) + u_t &= f(t), t \in \Omega = [0, 3\pi] \\ u(0) &= 0\end{aligned} \tag{31}$$

Let us define $f(t)$ to be given by

$$f(t) = \sin(\sin^2(t^2)) + 2t\cos(t^2) \tag{32}$$

and proceed by approximating $u(t)$ by a deep neural network.

Based on the analytical expression of the solution, it is convenient for error calculation of neural networks. It is worth noting that the analytical solution of this problem is the classic one-dimensional swept function $u(t) = \sin(t^2)$, so one can perform detailed calculations of the error of the different frequency components during the convergence process. As can be seen from Figure 2, the most significant difference between PICN and PINN in the convergence process is that PICN first learns the high-frequency components in the equation, while PINN first fits the low-frequency components. This phenomenon can be clearly seen from the frequency domain error distribution variation with the training epoch.

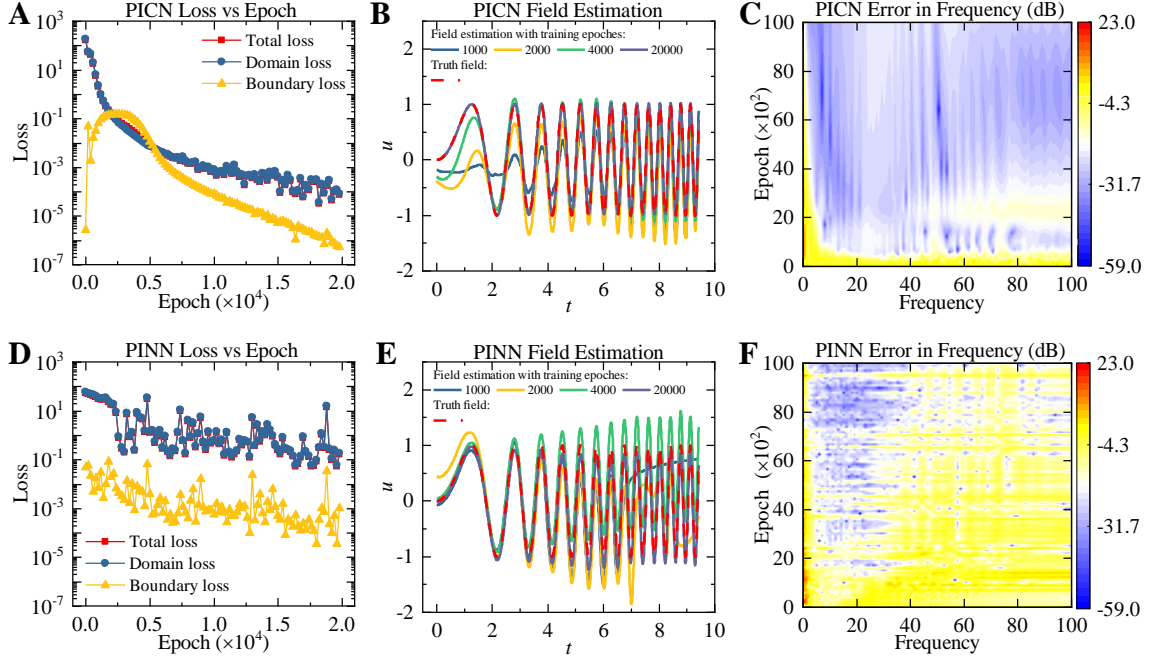

**Fig. 2** Comparison of the training process between PICN and PINN for solving a one-dimensional nonlinear differential equation. Loss function changes are compared in A and D; Estimation changes are compared in B and E; Normalized prediction errors in the frequency domain are compared in C and F. For PICN, the kernel size of DCNN is 1×1000, and the kernel size of CNN is 3x3. The PINN uses a 1-20-20-20-20-20-1 fully connected network. The training results are shown on 1000 points uniformly distributed in the defined domain. The weight ratio between governing loss and boundary loss is 9:1, the Adams optimizer is used, and the learning rate is 0.001. Training is performed on a Titan X GPU. It costs 40 seconds for PICN and 872s for PINN to train 20000 epochs.

Subsequently, we further solve the two-dimensional nonlinear partial differential equations containing trigonometric functions and square operators based on neural networks. The classical boundary value problem of the first kind is dealt with here. The following partial differential equation along with boundary conditions in a rectangular domain is given as follows,

$$\sin(u^2) + \Delta u = f(x, y) \quad (x, y) \in \Omega = [0, 10] \times [0, 6]$$
$$u(0, y) = \sin(y^2), u(10, y) = \sin(100 + y^2) \tag{33}$$
$$u(x, 0) = \sin(x^2), u(x, 6) = \sin(36 + x^2)$$

Let us define $f(x, y)$ to be given by

$$f(x, y) = \sin(\sin^2(x^2 + y^2)) + 4\cos(x^2 + y^2) - 4(x^2 + y^2)\sin(x^2 + y^2) \tag{34}$$

and proceed by approximating $u(x, y)$ by a deep neural network. The training parameters in the network can be learned by minimizing the error loss controlled by both boundary conditions and governing equations.

The analytical solution is $u(x, y) = \sin(x^2 + y^2)$. In this example, the prediction results of PICN and PINN show a more pronounced difference in the convergence behavior in the frequency domain. As shown in Fig. 3, from the comparison of Figure B and Figure D, PICN can first accurately estimate the high-frequency components of the result when the epoch is 500 and then accurately estimate the low-frequency components of the result. In addition, from Figure D, it can be seen that PINN encountered apparent obstacles in the process of estimating high-frequency components. This phenomenon did not appear in the training process of PICN. In panels C and D in Figure 4, the lower-left corner represents the low-frequency component of the normalized error, and the upper right corner represents the high-frequency content of the signal. It can be seen from the comparison that PICN also shows excellent learning ability for the high-frequency components of the signal, compared to PINN. The above analysis and comparison demonstrate the PICN potential advantages of approximating physical fields with multi-frequency components.

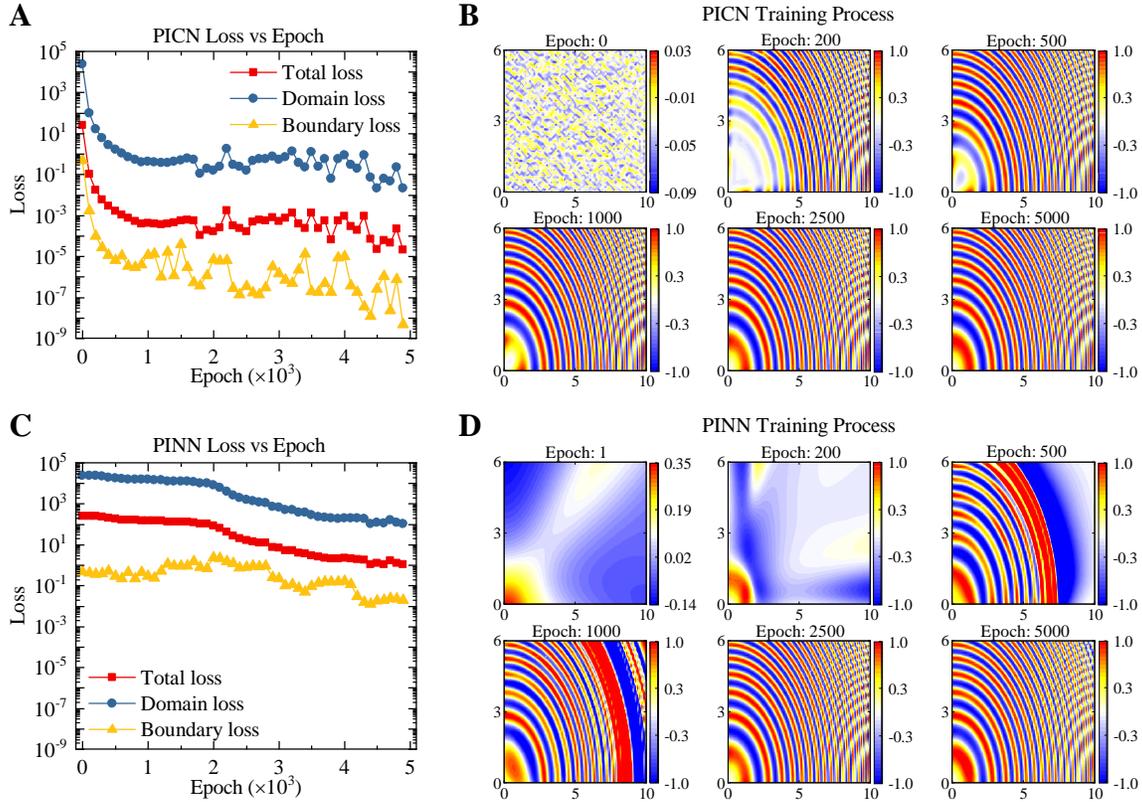

**Fig. 3** The comparison between the training process of PICN and PINN to solve a nonlinear partial differential equation boundary value problem. Loss function changes are compared in A and C; Estimation changes are compared in B and D. For PICN, the kernel size of DCNN is 120×200, and the kernel size of CNN is 3x3. The PINN uses a 1-20-20-20-20-20-1 fully connected network. The data for calculating governing loss come from 24000 points uniformly distributed in the defined domain. The data for calculating boundary loss comes from 636 points uniformly distributed on the boundary. The weight ratio between governing loss and boundary loss is 999:1 for PICN and 99:1 for PINN, the Adams optimizer is used, and the learning rate is 0.01 for PICN and 0.0001 for PINN. Training is performed on a Titan X GPU. It costs 61s for PICN to train 5000 epochs which is much less than that for PINN.

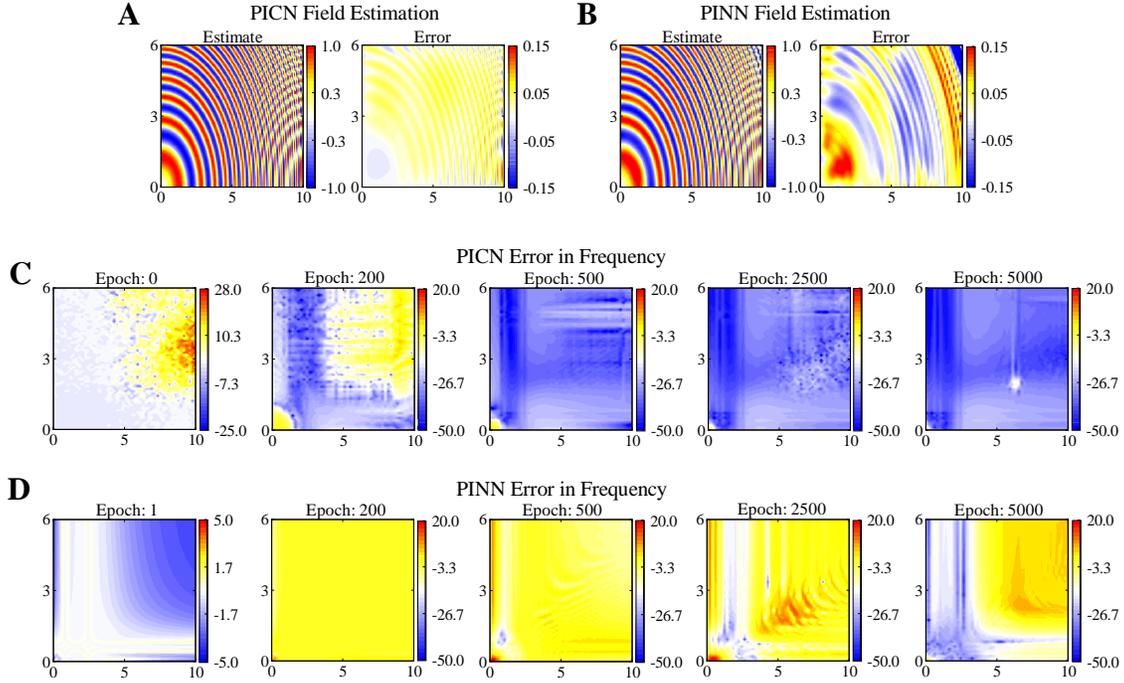

**Fig. 4** Model error comparisons. The final estimation error comparison is given by Figures A, and B. Normalized prediction errors in the frequency domain are compared in C and D.

## 4. Data-driven solver for nonlinear physical field on a regular domain

### 4.1. Example (ODE with sine nonlinearity)

This section shows the solving capability of PICN for some complex nonlinear differential equations. As an example, let us consider a nonlinear ordinary differential equation with a sine nonlinearity. In one dimension case, the following differential equations and their boundary conditions are adopted

$$u_x + \sin u_x + u(x) = q(x), x \in \Omega = [0,3]$$
$$u(0) = 0 \tag{35}$$

This problem has the exact analytical solution as

$$u(x) = \exp(-x)\sin(m\pi x^2) \tag{36}$$

Note that $q(x)$ can be found by direct substitution.

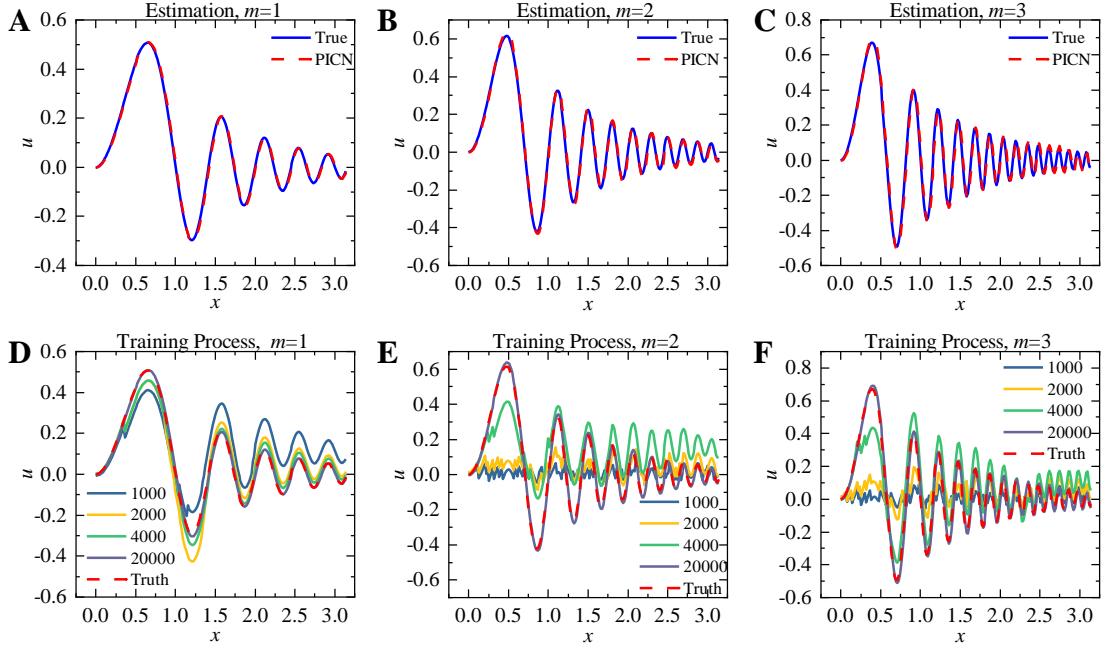

**Fig. 5** The training process of solving differential equations with sine nonlinearity using PICN. The first row shows the final estimation; the second row shows the convergence process. Parameters for the three cases are m=1, 2, and 3. For PICN, the kernel size of DCNN is 1×200, and the kernel size of CNN is 3×3. The training data to evaluate the governing loss contains 200 points uniformly distributed across the defined domain. The weight ratio between governing loss and boundary loss is 9:1, the Adams optimizer is used, and the learning rate is 0.01.

The training process for this problem is shown in detail in Figure 5. Note that only a single observation of the physical field is known in this problem, and the exact solution to this problem is characterized by complex fluctuations with multiple frequency components. The solution results of the shallow neural network proposed in this paper show that by comprehensively considering the governing equations determined by physical laws, one can accurately capture the complex solutions of nonlinear equations with only a small amount of boundary data.

### 4.2. Example (PDE involving Laplace and nonlinear sine-square operators)

This example emphasizes our method's ability to handle hybrid boundary conditions and is applicable to partial differential equation problems controlled jointly by linear and nonlinear operators. Laplacian operators appear in various areas of applied mathematics, including thermal, flow, and electric field analysis. Here, the following mixed boundary value problem of nonlinear partial differential equations with Laplace operator is analyzed

$$\sin(u^2) + \Delta u = f(x, y) \qquad (x, y) \in \Omega = [0, 5] \times [0, 3] \qquad (37)$$

where

$$f(x) = \sin(\sin^2(x^2 + y^2)) + 4\cos(x^2 + y^2) - 4(x^2 + y^2)\sin(x^2 + y^2) \qquad (38)$$

And the boundary conditions are

$$\begin{cases} \dfrac{\partial u}{\partial \mathbf{n}} = -2x\cos(x^2 + y^2) & x = 0, 0 \leq y \leq 3 \\ u(x, y) = \sin(x^2 + y^2) & \text{other boundaries} \end{cases} \qquad (39)$$

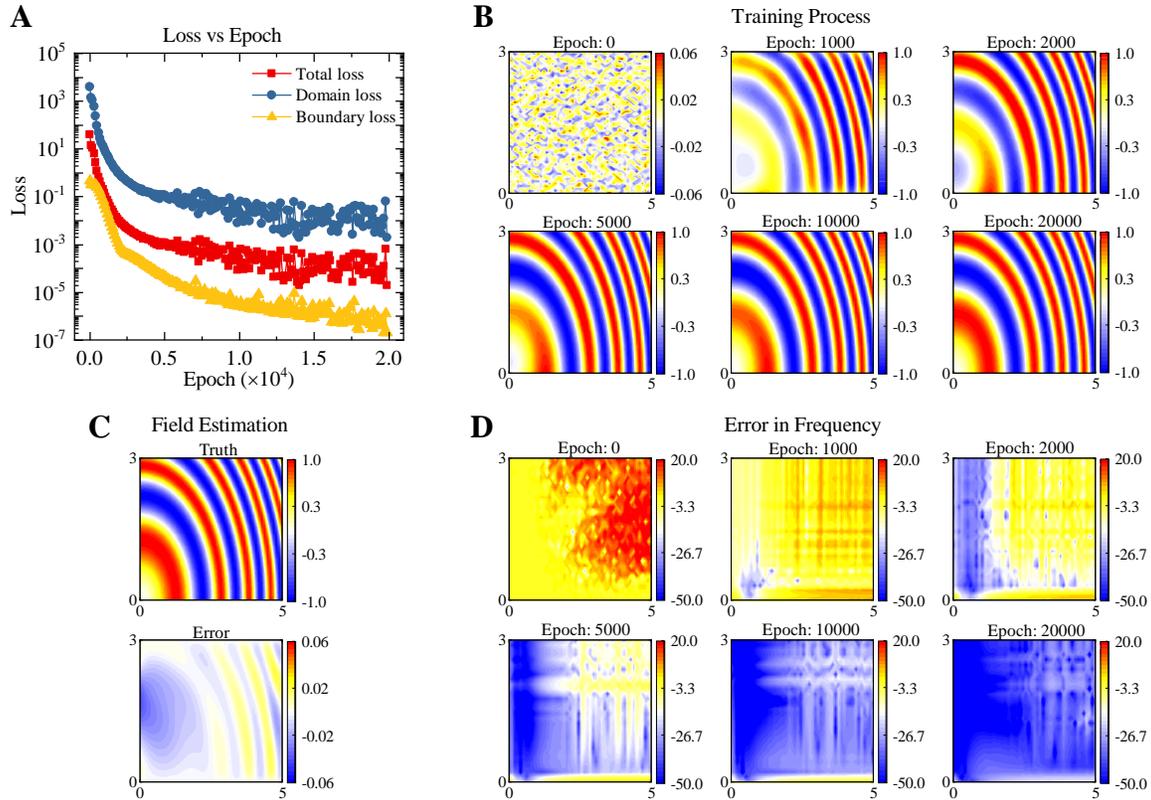

**Fig. 6** The training process of solving differential equations using PICN. Loss function changes are shown

in A; Estimation changes are shown in B. The final estimation error is given by C. Normalized prediction errors in the frequency domain are shown in D. The kernel size of DCNN is 60×100, and the kernel size of CNN is 3×3. The data for governing loss come from 6000 points uniformly distributed in the defined domain. The data for calculating boundary loss comes from 320 points uniformly distributed on the boundary. The weight ratio between governing loss and boundary loss is 99:1, the Adams optimizer is used, and the learning rate is 0.001.

The analytical solution is $u(x,y)=\sin(x^2+y^2)$. The solving process of this problem is shown in Figure 6. It can be seen that PICN can be directly used for solving mixed boundary conditions without any additional network structure changes or additional preprocessing frameworks. From the comparison with the example in Section 3.2, it can be seen that the influence of the mixed boundary conditions on the network convergence can be ignored, and the distribution of the network estimation error in the frequency domain remains consistent.

## 4.3. Example (Schrödinger equation)

The nonlinear Schrödinger equation is adopted to analyze the feasibility of neural network methods in dealing with governing partial differential equations with nonlinear operators and complex-valued solutions. This paper discusses the following time-dependent one-dimensional problem. The equation to be solved is:

$$i\psi_t + \psi_{xx} + \psi - |\psi|^2 \psi = 0 \quad (x,t) \in \Omega = [0,\pi] \times [0,\pi] \tag{40}$$

where $\psi = u + iv$.

Using the form of the wave function $\psi = \exp(i(x-t))$, the equation can be transformed as

$$\begin{cases} u_t + v_{xx} + v - (u^2+v^2)v = 0 \\ v_t - u_{xx} - u + (u^2+v^2)u = 0 \end{cases} \tag{41}$$

The analytical solution to this problem is $u=\cos(x-t), v=\sin(x-t)$. And a simple Dirichlet boundary condition is used here.

To demonstrate the effectiveness of the PICN method, the error between the exact solution and the prediction,

and the training results at different epochs are given in Figure 7 in detail. The example implementation means that, based on physical information machine learning, one can accurately analyze the complex nonlinear behavior governed by the Schrödinger equation using small-sample data on boundaries. Here, the Schrödinger Equation is transformed as the system of partial differential equations which can be directly solved in real-domain. This means that the PICN is capable of solving nonlinear partial differential equations. It is worth noting that for a system of nonlinear partial differential equations, the network structure, training data distribution, and optimizer of PICN do not need to be specifically customized and delicately optimized, which provides the possibility for coupled analysis of complex multiphysics problems.

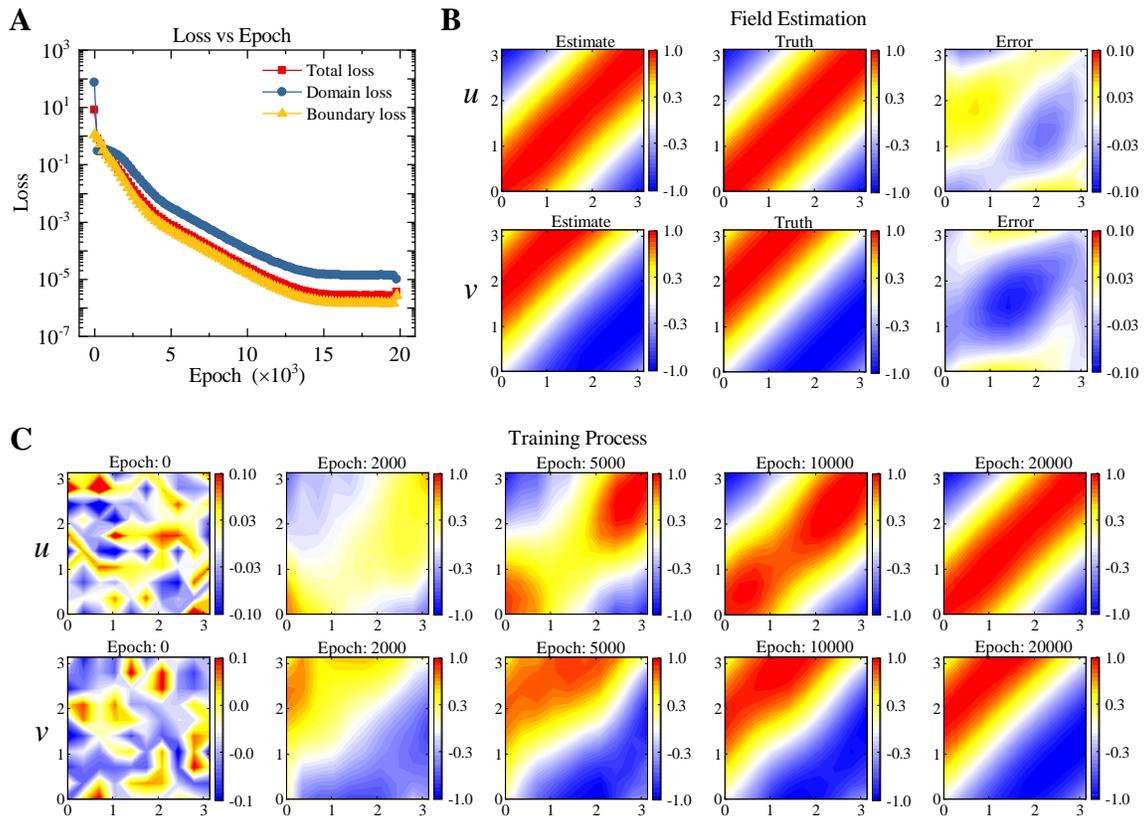

**Fig. 7** The training process of solving differential equations using PICN. Loss function changes are shown in A; The final estimation error is given by B; Estimation changes are shown in C. The kernel size of DCNN is 10×10, and the kernel size of CNN is 3x3. The data for calculating governing loss comes from 100 points uniformly distributed in the defined domain. The data for calculating boundary loss comes from 40 points

uniformly distributed on the boundary. The weight ratio between governing loss and boundary loss is 1:9, the Adams optimizer is used, and the learning rate is 0.001.

## 5. Data-driven solver for the nonlinear physical fields on irregular domains

### 5.1. Example (physical field on the star-shaped domain)

This example investigates a nonlinear PDE defined in a star-shaped domain, which domain is defined parametrically by:

$$(x,y) \in \Omega = \{\rho \leq 1 + \cos^2(4\theta)\} \tag{42}$$

in which $\theta \in [0, 2\pi]$, and $(\rho, \theta)$ is the polar coordinate of $(x, y)$ satisfying $(x, y) = (\rho\cos\theta, \rho\sin\theta)$.

Here, the equation to be solved is $u_x + uu_y = f(x)$, and the Dirichlet boundary condition is used. Let us define $f(x,t)$ to be given

$$f(x) = -\sin(x + kxy + ky^2)(ky+1) - \sin(x + kxy + ky^2)\cos(x + kxy + ky^2)(kx + 2ky) \tag{43}$$

The exact solution is $u(x,y) = \cos(x + kxy + ky^2)$.

Figure 8 shows the comparison of PICN prediction and the exact solution and the loss convergence process when $k$ =5. It can be seen that the predicted physical field using PICN can better capture the oscillating properties of the physical field on this irregular domain. Moreover, the PICN predictions are almost the same as the analytical results at most positions in the definition domain.

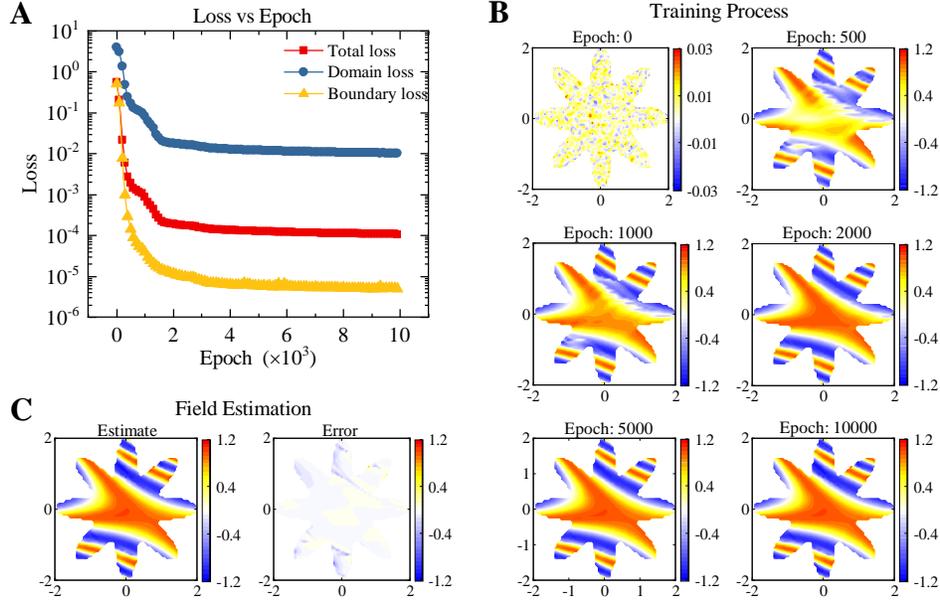

**Fig. 8** The training process of solving differential equations using PICN. Loss function changes are shown in A; Estimation changes are shown in B; The final estimation error is given by C. The kernel size of DCNN is 200×200, and the kernel size of CNN is 3×3. The training data comes from points uniformly distributed in the defined domain. The interval between training points is 0.02. The data for calculating the boundary loss comes from 800 points with uniform distribution of angles. The weight ratio between governing loss and boundary loss is 1:9, the Adams optimizer is used, and the learning rate is 0.001.

## 5.2. Example (physical field on a bird-like domain)

Consider the following bird-like 2D domain problem. The boundary of this bird-like domain is defined parametrically by

$$\partial \Omega = \{(\rho\cos\theta, \rho\sin\theta) : \rho = e^{\sin\theta}\sin^2(3\theta) + e^{\cos\theta}\cos^2(3\theta)\} \quad (44)$$

where $\theta \in [0, 2\pi]$, and $(\rho, \theta)$ is the polar coordinate of $(x, y)$ satisfying $(x, y) = (\rho\cos\theta, \rho\sin\theta)$.

The partial differential equation involving Laplace and sine-square operator in Section 4.2 is revisited, and Dirichlet data is known along the boundary. The exact solution is also $u(x, y) = \cos(kxy + ky)$. The differences between the PICN computation and the exact solution and training process of PICN are given in Figure 9.

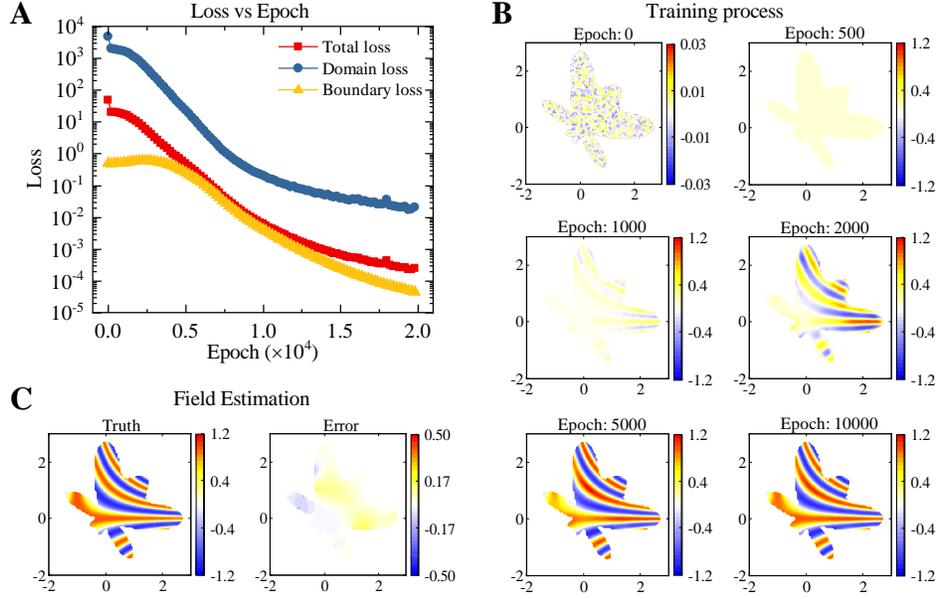

**Fig. 9** The training process of solving differential equations using PICN. Loss function changes are shown in A; Estimation changes are shown in B; The final estimation error is given by C. The kernel size of DCNN is 250×250, and the kernel size of CNN is 3x3. The training data comes from points uniformly distributed in the defined domain. The interval between training points is 0.02. The data for calculating the boundary loss comes from 800 points with uniform distribution of angles. The weight ratio between governing loss and boundary loss is 1:99, the Adams optimizer is used, and the learning rate is 0.0001

## 5.3. Example (physical field on a starfish domain)

The example on the star-shaped domain considers nonlinear combinations $uu_y$ of physical field and its first derivatives, and the example on the star-shaped domain involves the Laplace and sine-square operator. In the following, a new nonlinear operator equation is considered. The generalized Laplace operator $c_1 u_{xx} + c_2 u_{yy}$ is considered, where the ratio $c_1 / c_2$ represents the orthotropic properties in the physical laws. In addition, a complex nonlinear combination of $\sin(u)$ and $u_x$ is also considered.

The equation to be solved is:

$$u_{xx} + 5u_{yy} + k\sin(u)u_x = f(x, y) \tag{45}$$

$$f(x,y) = -\left(1 + 2ky + k^2\left(5x^2 + 20xy + 21y^2\right)\right)\sin\left[0.5 + x + kxy + ky^2\right]$$
$$+ k\cos\left(0.5 + x + kxy + ky^2\right)\left(10 + (1 + ky)\sin\left[\sin\left((0.5 + x + kxy + ky^2)\right)\right]\right) \quad (46)$$

Here, a boundary of the starfish domain is employed, which can be expressed parametrically by

$$\partial\Omega = \{(\rho\cos\theta, \rho\sin\theta) : \rho = 1 + 0.5\cos^2(2.5\theta)\} \quad (47)$$

where $\theta \in [0, 2\pi]$, and $(\rho, \theta)$ is the polar coordinate of $(x, y)$.

The solution is $u(x, y) = \sin(0.5 + x + kxy + ky^2)$, and Dirichlet boundaries are still used in the PICN analysis under $k=5$. As shown in Figure 10, this PICN prediction result is still to demonstrate the use of PICN to solve the irregular domain problem. Of course, when the simple linear interpolation network is used for boundary processing in this paper, this algorithm only ensures the first-order algebraic accuracy on the boundary. Some delicate technologies such as complex high-order interpolation or interpolation networks trained with hybrid information of each order of derivatives can be introduced to ensure high numerical precision for irregular boundary problems.

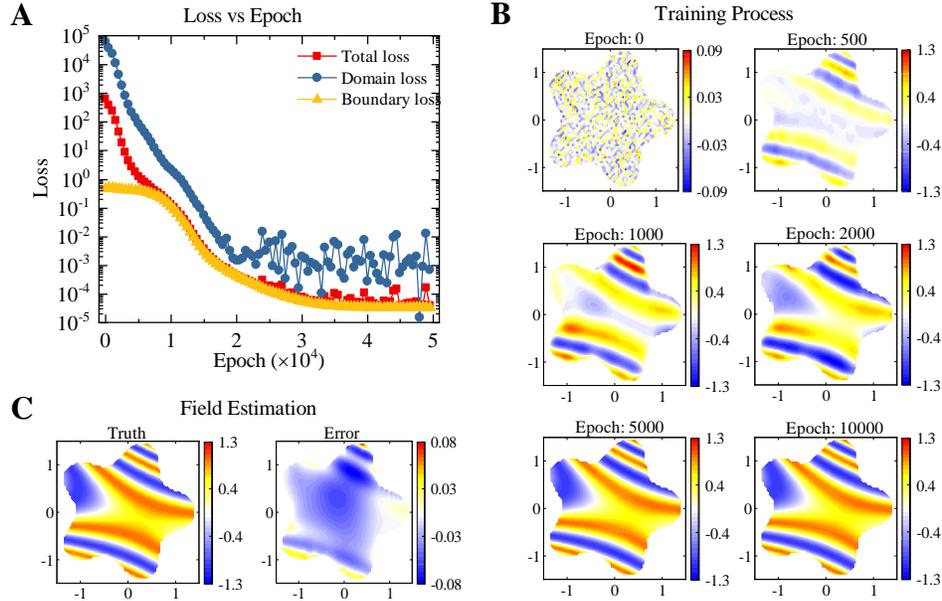

**Fig. 10** The training process of solving differential equations using PICN. Loss function changes are shown in A; Estimation changes are shown in B; The final estimation error is given by C. The kernel size of DCNN

is 60×60, and the kernel size of CNN is 3×3. The training data comes from points uniformly distributed in the defined domain. The interval between training points is 0.02. The data for calculating the boundary loss come from 800 points with uniform distribution of angles. The weight ratio between governing loss and boundary loss is 1:99, the Adams optimizer is used, and the learning rate is 0.0001

## 6. Learning physical field from noisy data

### 6.1. Data-driven discovery of physical law

This part presents the data-driven discovery of physical law using PICN networks. The data-driven discovery corresponds to estimating the parameters in the physical laws. First, take $\lambda_1 u_{xx} + \lambda_2 u_{yy} = 0$ as an example, which corresponds to the case of $\mathbf{T}[u] = \lambda_1 u_{xx} + \lambda_2 u_{yy}, \lambda = (\lambda_1, \lambda_2)$. This is a typical heat conduction problem of orthotropic materials. It is necessary to estimate the parameters based on some observation data that may contain noise and determine the optimal parameters $\lambda$ to characterize this physical problem. This essentially corresponds to how to accurately determine the heat conduction parameters of the material in the two orthogonal directions of $x$ and $y$ from the temperature field observed in the experiment.

From a finite element analysis under a specific boundary condition, the finite element results and random noise are added to replace the experimental observations of the temperature field. And then, PICN machine learning is executed on the experimental observations. Finally, the anisotropic heat conduction parameter ratio of the material is estimated. Figure 11 shows the PICN estimation results based on noise-free experimental observations, which can achieve high-precision temperature field reconstruction consistent with experimental observations. It is worth noting that the interpolation network is trainable, and its network weights just correspond to the parameters $\lambda$ in the operator T. From the interpolation network weights in the trained PICN, we get an estimate of the parameters that satisfy $\lambda_2 / \lambda_1 = 5.0017323$, which is completely consistent with the value of 5 that used to construct our physical field observations.

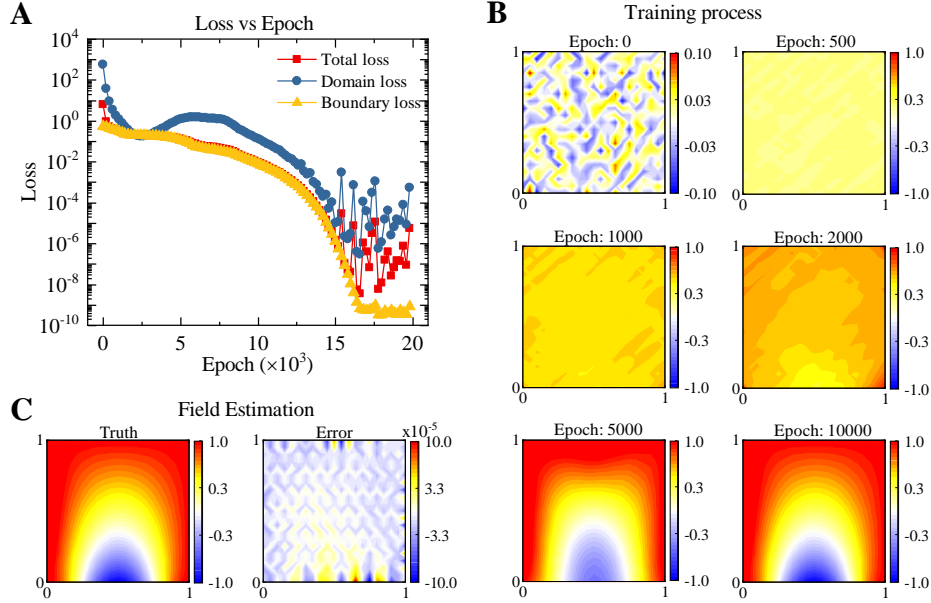

**Fig. 11** The training process of PICN to discover nonlinear field equations. Loss function changes are shown in A; Estimation changes are shown in B; The final estimation error is given by C. The kernel size of DCNN is 20×20, and the kernel size of CNN is 3×3. The training data comes from points uniformly distributed in the defined domain. The interval between training points is 0.05. The weight ratio between governing loss and boundary loss is 1:99, the Adams optimizer is used, and the learning rate is 0.0002

To further examine the capabilities of this PICN machine learning, we conduct a comparative study of the problem under different noise corruption levels. The case of experimental observations affected by Gaussian noise of different degrees is further discussed in Figure 12. It can be seen that, although the estimation error increases correspondingly as the noise level increases, the anisotropic ratio $\lambda_2/\lambda_1$ can be estimated fairly accurately from PICN Learning experimental observations with the Gaussian noise when the mean is 0 and the standard deviation below 0.1. This shows that the PICN has strong robustness for the parameter evaluation, and even if the standard deviation reaches 0.1, a reasonably good parameter evaluation can be produced from the present PICN.

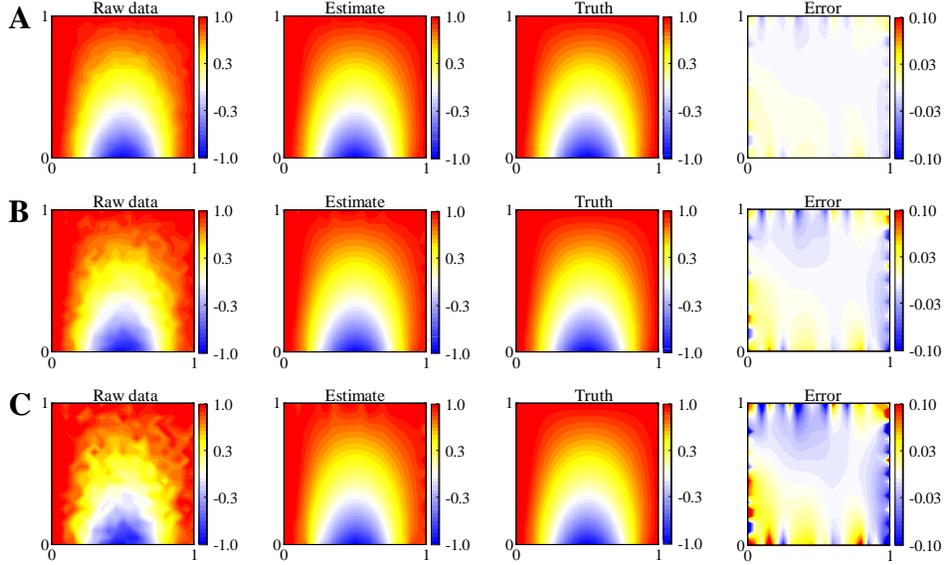

**Fig. 12** Parameter estimation of partial differential equations under different noise levels. The case where the standard deviation of the Gaussian noise is 0.02 is shown in A, the case where the standard deviation of the Gaussian noise is 0.05 is shown in B, and the case where the standard deviation of the Gaussian noise is 0.1 is shown in C.

**Table 1.** Estimation of anisotropic ratio $\lambda_2 / \lambda_1$ from PICN learning under different Gaussian noise

| Standard deviation of Noise | 0 | 0.02 | 0.05 | 0.10 |
|---|---|---|---|---|
| **Anisotropic ratio estimation** | 5.0017 | 5.0114 | 5.0379 | 5.1218 |
| **Error** | 0.034% | 0.23% | 0.76% | 2.44% |

## 6.2. Denoising display with physical information

In addition, the data-driven discovery also corresponds to a denoising display of a noisy observation. The generation scheme of observation data mentioned above is still adopted to implement and analyze this problem. In this problem, the physical information is the temperature field measurement of an orthotropic square plate with known material parameters. This confirms that the physical field is governed by the anisotropic heat conduction equation, which is expressed as $\lambda_1 u_{xx} + \lambda_2 u_{yy} = 0$. When PICN is used to train the loss jointly controlled by the observation and the physical information, the training result can just achieve the denoising display of the physical field. As can be seen from Figure 13, when the form of the physical governing equations is known, PICN can accurately generate a more clear physical display from noisy sampled data. Even if the standard deviation of the error is as high as 0.4, the error between the re-established results and ideal observations is still only about 0.1.

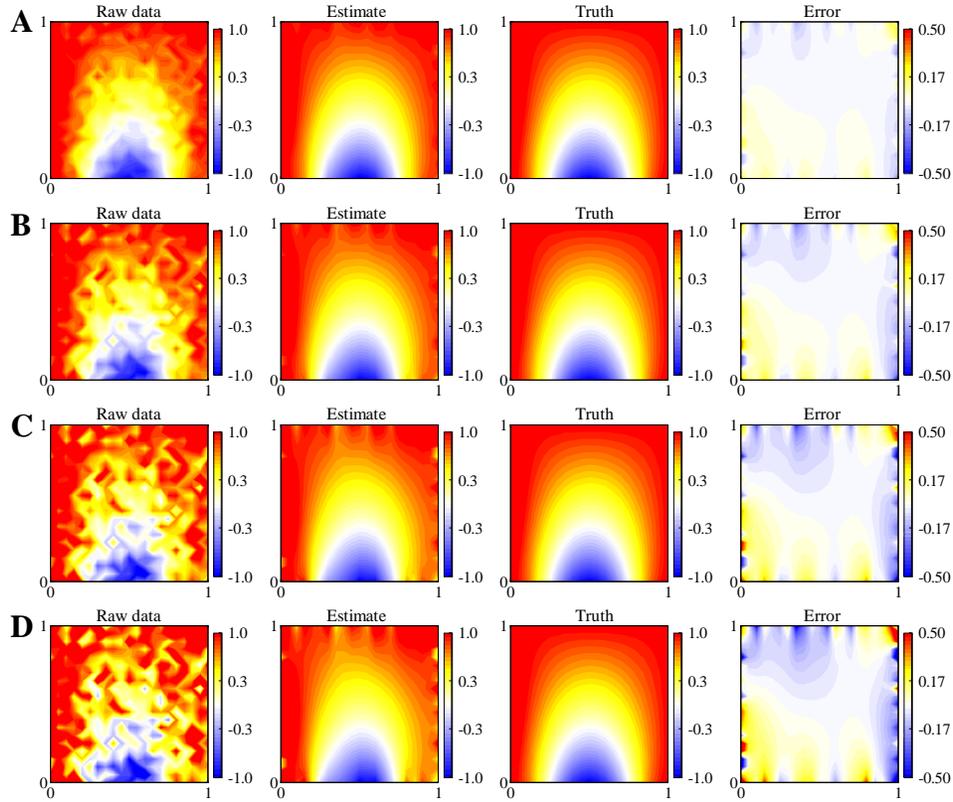

**Fig. 13** Denoising display with physical information under different noise conditions. The case where the standard deviation of the Gaussian noise is 0.1 is shown in A, the case where the standard deviation of the Gaussian noise is 0.2 is shown in B, the case where the standard deviation of the Gaussian noise is 0.3 is shown in C, and the case where the standard deviation of the Gaussian noise is 0.4 is shown in D.

Even when the governing equations satisfied by the physics are not precisely known, the denoising display of the physics can be achieved programmatically. Here, a physical field in the explicit functional form is used for the analysis by $f(x,y)=\sin(x+5y)+\exp(-x)$. The interpolation network with trainable weights is still used, and then the noisy physics observations are learned and trained based on PINN. Here, the equation $k_1 u_{xx} + k_2 u_{yy} = 1$ is used to describe the physical field. As shown in Figure 14, the machine learning results show that PICN achieves denoising by using enhancements of physics information. It can be seen that even if the equation of the original physical field is unknown, a smooth approximation of the physical field can be obtained by choosing a simple governing equation that may not be fully applicable to the physical field. The results show that this PICN scheme still shows good robustness for the case where the standard deviation of the Gaussian noise is 0.3.

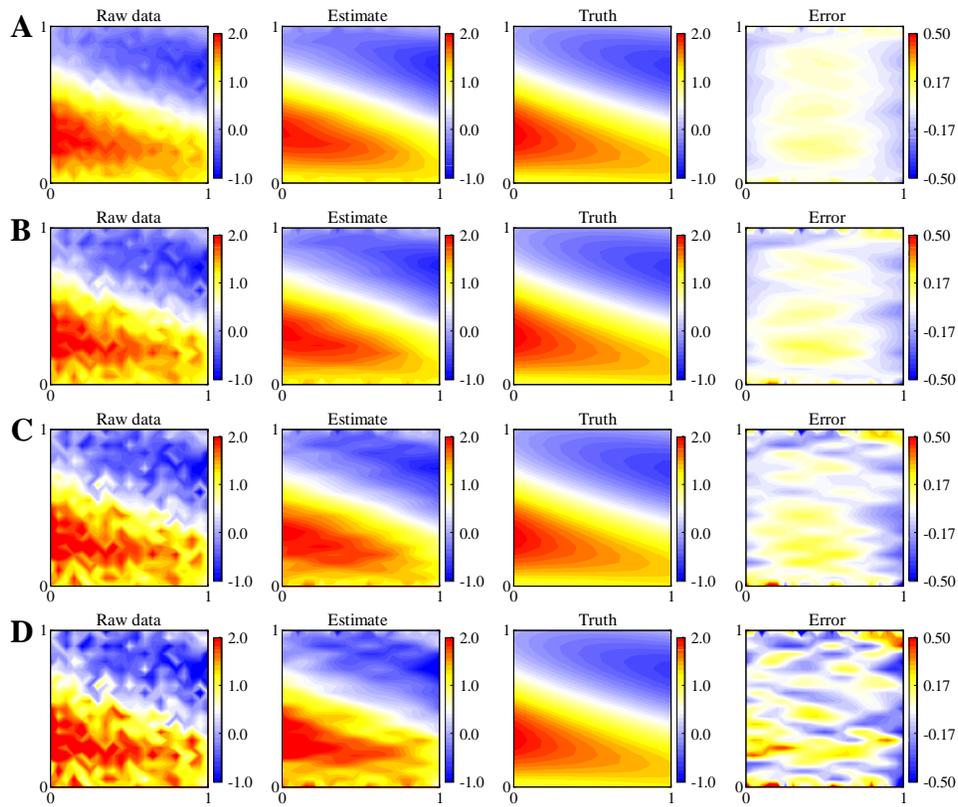

**Fig. 14** Denoising display without physical information under different noise conditions. The case where the standard deviation of the Gaussian noise is 0.1 is shown in A, the case where the standard deviation of the Gaussian noise is 0.2 is shown in B, the case where the standard deviation of the Gaussian noise is 0.3 is shown in C, and the case where the standard deviation of the Gaussian noise is 0.3 is shown in D.

## 7. Conclusions

When using the fully connected network (FCN) backend PINN method to estimate the physical field with multi-frequency components, the spectral bias nature may lead to excessive training overhead and even does not ensure the convergence of the training process. This paper recommends a physical information convolutional network (PICN) solver in physics-informed machine learning from a CNN perspective. The potential advantages of the PINN network are discussed theoretically and numerically on approximating physical fields with multi-frequency components. Further, a series of numerical experimental cases demonstrate the feasibility of PINN in two types of physics-informed machine learning problems containing data-driven solutions and data-driven discovery. Based on the CNN perspective, this network solver provides a new alternative solution for physics-informed machine learning.

## Declaration of competing interest

The authors declare that they have no known competing financial interests or personal relationships that could have appeared to influence the work reported in this paper.

## Acknowledgments

This research is supported by financial support from the National Natural Science Foundation of China (NNSFC) (Grant Nos. 11802225 and 61805185).

## References

[1]   M. I. Jordan and T. M. Mitchell, "Machine learning: Trends, perspectives, and prospects," *Science*, vol. 349, no. 6245, pp. 255–260, 2015.

[2]   I. Arel, D. C. Rose, and T. P. Karnowski, "Deep machine learning-a new frontier in artificial intelligence research," *IEEE computational intelligence magazine*, vol. 5, no. 4, pp. 13–18, 2010.

[3]   Ibrahim Kök, M. U. Şimşek, and S. Özdemir, "A deep learning model for air quality prediction in smart cities," in *2017 IEEE International Conference on Big Data*, 2017, pp. 1983–1990.

[4]  N. Sünderhauf et al., "The limits and potentials of deep learning for robotics," *The International Journal of Robotics Research*, vol. 37, no. 4–5, pp. 405–420, 2018.

[5]  U. Hasson, S. A. Nastase, and A. Goldstein, "Direct fit to nature: An evolutionary perspective on biological and artificial neural networks," *Neuron*, vol. 105, no. 3, pp. 416–434, 2020.

[6]  B. Norgeot, B. S. Glicksberg, and A. J. Butte, "A call for deep-learning healthcare," *Nature medicine*, vol. 25, no. 1, pp. 14–15, 2019.

[7]  P. Schneider et al., "Rethinking drug design in the artificial intelligence era," *Nature Reviews Drug Discovery*, vol. 19, no. 5, pp. 353–364, 2020.

[8]  L. Bottou, F. E. Curtis, and J. Nocedal, "Optimization methods for large-scale machine learning," *Siam Review*, vol. 60, no. 2, pp. 223–311, 2018.

[9]  R. H. Hariri, E. M. Fredericks, and K. M. Bowers, "Uncertainty in big data analytics: survey, opportunities, and challenges," *Journal of Big Data*, vol. 6, no. 1, pp. 1–16, 2019.

[10] C. Tan, F. Sun, T. Kong, W. Zhang, C. Yang, and C. Liu, "A survey on deep transfer learning," in *International Conference on Artificial Neural Networks*, 2018, pp. 270–279.

[11] N. Huang, M. Slaney, and M. Elhilali, "Connecting deep neural networks to physical, perceptual, and electrophysiological auditory signals," *Frontiers in neuroscience*, vol. 12, pp. 532, 2018.

[12] X. Hu, H. Hu, S. Verma, and Z.-L. Zhang, "Physics-guided deep neural networks for power flow analysis," *IEEE Transactions on Power Systems*, vol. 36, no. 3, pp. 2082–2092, 2020.

[13] M. A. Nabian and H. Meidani, "Physics-driven regularization of deep neural networks for enhanced engineering design and analysis," *Journal of Computing and Information Science in Engineering*, vol. 20, no. 1, p. 11006, 2020.

[14] N. Borodinov, S. Neumayer, S. v Kalinin, O. S. Ovchinnikova, R. K. Vasudevan, and S. Jesse, "Deep neural networks for understanding noisy data applied to physical property extraction in scanning probe microscopy," *NPJ Computational Materials*, vol. 5, no. 1, pp. 1–8, 2019.

[15] Z. Fang, "A High-Efficient Hybrid Physics-Informed Neural Networks Based on Convolutional Neural Network," *IEEE Transactions on Neural Networks and Learning Systems*, pp. 1–13, 2021.

[16] M. Raissi, P. Perdikaris, and G. E. Karniadakis, "Physics Informed Deep Learning (Part I): Data-driven Solutions of Nonlinear Partial Differential Equations," *Journal of Computational Physics*, vol. 378, no. Part I, pp. 1–22, 2019.

[17] M. Raissi, A. Yazdani, and G. E. Karniadakis, "Hidden fluid mechanics: Learning velocity and pressure fields from flow visualizations," *Science*, vol. 367, no. 6481, pp. 1026–1030, 2020.

[18] G. E. Karniadakis, I. G. Kevrekidis, L. Lu, P. Perdikaris, S. Wang, and L. Yang, "Physics-informed machine learning," *Nature Reviews Physics*, vol. 3, no. 6, pp. 422–440, 2021.

[19] J.-L. Wu, H. Xiao, and E. Paterson, "Physics-informed machine learning approach for augmenting turbulence models: A comprehensive framework," *Physical Review Fluids*, vol. 3, no. 7, p. 74602, 2018.

[20] E. Qian, B. Kramer, B. Peherstorfer, and K. Willcox, "Lift & learn: Physics-informed machine learning for large-scale nonlinear dynamical systems," *Physica D: Nonlinear Phenomena*, vol. 406, p. 132401, 2020.

[21] J.-X. Wang, J.-L. Wu, and H. Xiao, "Physics-informed machine learning approach for reconstructing Reynolds stress modeling discrepancies based on DNS data," *Physical Review Fluids*, vol. 2, no. 3, p. 34603, 2017.

[22] K. Kashinath *et al.*, "Physics-informed machine learning: case studies for weather and climate modelling,"

*Philosophical Transactions of the Royal Society A*, vol. 379, no. 2194, p. 20200093, 2021.

[23]  G. Pilania, K. J. McClellan, C. R. Stanek, and B. P. Uberuaga, "Physics-informed machine learning for inorganic scintillator discovery," *The Journal of chemical physics*, vol. 148, no. 24, p. 241729, 2018.

[24]  D. R. Harp, D. O'Malley, B. Yan, and R. Pawar, "On the feasibility of using physics-informed machine learning for underground reservoir pressure management," *Expert Systems with Applications*, vol. 178, p. 115006, 2021.

[25]  D. Zhang, L. Lu, L. Guo, and G. E. Karniadakis, "Quantifying total uncertainty in physics-informed neural networks for solving forward and inverse stochastic problems," *Journal of Computational Physics*, vol. 397, p. 108850, 2019.

[26]  N. Rahaman *et al.*, "On the spectral bias of neural networks," in *International Conference on Machine Learning*, 2019, pp. 5301–5310.

[27]  Z.-Q. J. Xu, "Frequency Principle: Fourier Analysis Sheds Light on Deep Neural Networks," *Communications in Computational Physics*, vol. 28, no. 5, pp. 1746–1767, 2020.

[28]  A. Veit, M. J. Wilber, and S. Belongie, "Residual networks behave like ensembles of relatively shallow networks," *Advances in neural information processing systems*, vol. 29, pp. 550–558, 2016.

[29]  J. Pan, E. Sayrol, X. Giro-i-Nieto, K. McGuinness, and N. E. O'Connor, "Shallow and deep convolutional networks for saliency prediction," in *Proceedings of the IEEE conference on computer vision and pattern recognition*, 2016, pp. 598–606.

[30]  N. B. Erichson, L. Mathelin, Z. Yao, S. L. Brunton, M. W. Mahoney, and J. N. Kutz, "Shallow neural networks for fluid flow reconstruction with limited sensors," *Proceedings of the Royal Society A*, vol. 476, no. 2238, p. 20200097, 2020.

[31] A. Creswell, T. White, V. Dumoulin, K. Arulkumaran, B. Sengupta, and A. A. Bharath, "Generative adversarial networks: An overview," *IEEE Signal Processing Magazine*, vol. 35, no. 1, pp. 53–65, 2018.

[32] Z. Li *et al.*, "Fourier neural operator for parametric partial differential equations," *arXiv preprint arXiv:2010.08895*, 2020.

[33] A. T. Mohan, N. Lubbers, D. Livescu, and M. Chertkov, "Embedding hard physical constraints in convolutional neural networks for 3D turbulence," *ICLR 2020 Workshop on Integration of Deep Neural Models and Differential Equations*, vol. 0, no. 2015, pp. 1–17, 2020.

[34] M. Abadi *et al.*, "Tensorflow: A system for large-scale machine learning," in *12th USENIX symposium on operating systems design and implementation*, 2016, pp. 265–283.

[35] G. D. Smith, G. D. Smith, and G. D. S. Smith, *Numerical solution of partial differential equations: finite difference methods*. Oxford university press, 1985.